# Evidential Reasoning in a Categorial Perspective: Conjunction and Disjunction of Belief Functions

Robert Kennes[1]
IRIDIA, Université Libre de Bruxelles
Av. F. D. Roosevelt 50 – CP 194/6
B-1050 Brussels, Belgium

## Abstract

The *categorial approach* to evidential reasoning can be seen as a combination of the *probability kinematics* approach of Richard Jeffrey (1965) and the *maximum (cross-) entropy inference* approach of E. T. Jaynes (1957). As a consequence of that viewpoint, it is well known that category theory provides natural definitions for logical connectives. In particular, disjunction and conjunction are modelled by general categorial constructions known as *products* and *coproducts*. In this paper, I focus mainly on Dempster-Shafer theory of belief functions for which I introduce a category I call *Dempster's category*. I prove the existence of and give explicit formulas for conjunction and disjunction in the subcategory of separable belief functions. In Dempster's category, the new defined conjunction can be seen as the *most cautious* conjunction of beliefs, and thus no assumption about *distinctness* (of the sources) of beliefs is needed as opposed to Dempster's rule of combination, which calls for *distinctness* (of the sources) of beliefs.

## 0 INTRODUCTION

J. Halpern and R. Fagin have pointed out [90 p.102] that belief functions can be understood in *'two useful and quite different ways ... The first as a generalized probability ... The second as a way of representing evidence ...*(i.e.) *as a mapping from probability functions to probability functions'*. This can be interpreted by saying that a belief function can be seen either as a static object (i.e., as a state of mind) or as a dynamic entity (i.e., as an evidence transforming a state of mind into another state of mind). The idea of putting together a static component with a dynamic one is not at all a new idea (cf. [Horvitz, Heckerman 86] for a nice survey of this idea). In 1965, Richard Jeffrey coined the term *probability kinematics* to emphasize the idea, although *probability dynamics* would have been a better term.

Actually, there exists a basic and simple mathematical structure encompassing the two foregoing views of belief functions: the *category* structure [Mac Lane 71]. The link between evidential reasoning and category theory can be best summarized by the so-called *Meseguer-Montanari correspondence* [Marti-Oliet, Meseguer 89]:

$$\text{States} \longleftrightarrow \text{Objects}$$
$$\text{Transitions} \longleftrightarrow \text{Morphisms}$$

where the set of states is a set of admissible *belief states* (or *opinions*) and the transitions from one belief state to a second belief state are the elements of a set of admissible *updatings* (or *adjustment of opinions*) transforming the first state into the second one.

I see two reasons for adopting a categorial viewpoint about evidential reasoning. The first one is related to the *probability kinematics* viewpoint, and the second one is related the *maximum entropy inference* approach to evidential reasoning. Actually, a category can be seen as an abstract view of a dynamic system of beliefs according to the probability kinematics viewpoint. An abstract point of view about beliefs has been advocated among others by Domotor [85]. Although the viewpoint adopted by Domotor is different from ours, the kind of structure he used – a monoid[2] of evidence operating[3] on a set of belief states, which is essentially the abstract view of a *machine* – always determines, in a natural way, a category. The second and more important reason for adopting a categorial viewpoint is: a very powerful way of defining objects is by using *universal properties* which is a generalization of defining objects by *maximum entropy* methods. *Defining an object by a universal property* is the categorist's way to *defining an object by a maximum (or minimum) principle*. Any mathematical objects defined by such a maximum principle live in a category. I also should mention that there already exits a maximum (or minimum) property principle known as *Principle of Minimum of Specificity* [Dubois, Prade 87a, 87b], *Principle of Minimum Commitment* [Hsia 91], or *Principle of Maximum Plausibility* [Smets 91] playing an increasing role in Dempster-Shafer theory. Nevertheless, the principle I will be using is not the former one and could be called *the least (or most) updated principle*.

---

[2] A monoid is a set along with a binary operation that is associative and has a neutral element 1.

[3] A monoid (M,*) operates on a set S iff every element m of M determines a transformation $\underline{m}$ of the set S such that for every elements m,n of M and for every element s of S: $(\underline{m*n})(s) = \underline{m}(\underline{n}(s))$ and $\underline{1}(s) = s$

The present paper is structured as following. Section 1 gives a detailed definition of categories. Section 2 presents the Boolean, the Bayesian and the Dempsterian categories of beliefs. Section 3 shows how conjunction and disjunction can be defined by using a maximum (or minimum) principle. Section 4 gives the definition of coproduct and conjunction in any category. Then, section 5 presents the product and disjunction. Finally, in section 6, I study the conjunction and disjunction of separable beliefs functions.

The product (disjunction) and coproduct (conjunction) of beliefs can be considered as an answer to questions raised by P.W. Williams [78, p.383] in his review of Shafer's book [76].

Some categorial approaches to probability theory and to evidential reasoning have already been proposed. Let us just mention F.W. Lawvere [Giry 82], Negoita [85], Goodman and Nguyen [85], Gärdenfors [88]. Contrary to evidential reasoning, the categorial study of fuzzy sets is now – since the work of J. Goguen [69] – a well established part of fuzzy set theory.

The problem of combining non-distinct experts opinions has already been examined in several papers, among them: [Smets 86], [Dubois, Prade 87b], [Hummel, Manevitz 87], [Ling, Rudd 89a,89b], [Wong, Lingras 90] and [Hau, Kashyap 90].

As the categorial framework is not very common among people concerned with evidential reasoning, this paper will be more expository than technical. Only a few technical results will be given and only very elementary notions of category theory will be presented and applied to evidential reasoning.

## 1 FROM THE DYNAMICS OF BELIEFS TO CATEGORIES OR ... VICE VERSA

The present section can be considered as an introduction to the idea of *category* for belief-minded people or as an introduction to the idea of *probability kinematics* for category-minded people. As a category is essentially a graph-theoretical structure I first need to give a precise definition of a graph (in fact of a directed multigraph):

**Definition**: a graph is defined by the following data:
1. a pair of classes P and A (whose elements are respectively called *points* and *arrows* or *objects* and *morphisms* or ... *states* and *transitions* ... according to our motivations)
2. together with a pair of maps s,t: A → P (the maps s and t are called *source* and *target* or *origin* and *extremity* or ... *initial state* and *final state*).

Some comments are worth mentioning:
1. The class P can be seen as the *static* component, whereas the class A can be seen as the *dynamic* component of the graph structure.
2. The static and dynamic components are linked together by the maps s and t, specifying the *initial* and *final* state of each arrow.
3. f: a → b means that f is an arrow whose source or initial state is a and whose target or final state is b. That is: f∈ A and s(f)=a and t(f)=b.
4. All arrows are directed: they all have an initial point and a final point. (*directed* graph)
5. Many arrows can share the same initial state and the same final state. (*multi* graph)

**Definition**:
a **category** is defined by the following **data**:
1. a graph s,t: A → P, together with:
2. a map i: P → A : a → $i_a$ . The arrow $i_a$: a → a is called the *identity arrow* at the point a
3. a partial map, called *composition*,
c: A×A → A: (f,g) → c(f,g)=f.g, f.g is called the *composite* of the arrows f and g (f.g is written in diagrammatic order), f.g is defined <u>iff</u> the target of f equals the source of g.

The preceding data must satisfy the following two **axioms**: for all points a,b and c and all arrows
f: a → b, g: b → c and h: c → d
(i) (f.g).h = f.(g.h)           (associativity of composition)
(ii) $i_a$.f = f = f.$i_b$      (identities are neutral for composition)

Intuitively, a category is simply a (directed multi-) graph together with a composition rule for *queueing* arrows satisfying associativity and with a neutral arrow at each point.

Examples: Many examples of categories can be classified according to some correspondences.
1. The major examples and motivations at the origins of category theory (in the 40's) were dominated by the following correspondence which could be called the *Klein correspondence* (after the famous Felix Klein 1872 Erlangen Program) :

*Structures*   ⟷   *Objects*
*Representations* ⟷ *Morphisms*

It is the Klein correspondence which has popularized the view of category as a meta-structure. But, I hasten to add that it is not the only possible view of categories.
Just two very classical examples:
(1) **SET** is the category whose objects are the sets, whose arrows are the usual functions, and the composition is the usual composition of functions. Instead of taking functions as arrows I could as well take the relations or the partial functions as arrows (in that case we are, of course, getting different categories).
(2) **RVECT** is the category whose objects are the real vector spaces, whose arrows are the linear mappings, and the composition is the usual composition of mappings.

In fact, any kind of structure together with a suitable notion of (homo)morphism give rise straightforward to a category.

2. In the late sixties appeared the *Lambek-Lawvere correspondence*:

$$Formulas \longleftrightarrow Objects$$
$$Proofs \longleftrightarrow Morphisms$$

also leading to categories. See [Marti-Oliet, Meseguer 89]

3. Taking into account with [Garvey, Lowrance, Fischler 81], [Hsia 90] and [Provan 90] that a belief function can be considered as a generalized formula and that an updating can be considered as a generalized proof (cf.[Pearl 88,90]), then we get the next correspondence:

$$Belief\ States \longleftrightarrow Objects$$
$$Updating \longleftrightarrow Morphisms$$

This can also be seen as a particular case of the *Meseguer-Montanari correspondence* (about concurrent systems) which appeared in the late eighties [Marti-Oliet, Meseguer 89]:

$$States \longleftrightarrow Objects$$
$$Transitions \longleftrightarrow Morphisms$$

In particular that correspondence associates a category to any machine.

4. Many important examples of categories do not fit into the preceding correspondences. A category can also be viewed as a common generalization of an algebraic structure: *the monoids*, and of an ordered structure: *the preordered sets*[4]. Explicitly, any monoid can be viewed as a category with only one object (take any object you want), the arrows are the elements of the monoid. Composition is the binary operation of the monoid. Any preordered set can be viewed as a category. The objects are the elements of the set and the arrows are the ordered pairs (a,b) of the preorder. In particular any ordered set[5] and any lattice[6] are (or can be viewed as) categories. Every monoid operating on a set (such an operation can be viewed as a machine) gives rise to a category in the following way: the objects are the elements of the set, and the arrows are the triples (a,m,b) where a and b are elements of the set and m is an element of the monoid such that m(a)=b. So, Domotor's [80] viewpoint is embedded is the categorial viewpoint.

## 2  CATEGORIES OF "BELIEFS"

I want to show here that there exist a lot of categories of "beliefs". That is, categories whose objects can be thought of as representing belief states (or opinions) of a cognitive agent concerning a particular situation. The general idea is the following one: If I adopt an abstract viewpoint of what should be a system of beliefs (of a cognitive agent), concerning a particular situation, I find natural to:

(i) first, consider a set of admissible *belief states* – whatever this term actually means –which can be taken by an agent, concerning the specific situation at hand.

(ii) second, consider a set of admissible *updatings* – whatever this term actually means – (determined by some evidence), transforming a belief state into another belief state.

(iii) third, the composition of two updatings should be an updating,

(iv) for each belief state there should exist a *trivial* updating, i.e., the one *doing nothing*.

We will get a category of beliefs each time we make clear each of the above notions which have been left vague. Here are some major examples:

The Boolean category of beliefs (induced from a special case of *Boolean machines* of [Domotor 80, p. 391] by the Meseguer-Montanari correspondence).

Let us consider a (finite or infinite) set $\Omega$ which can be interpreted as a set of possible values for a variable, or possible answers to a question. The boolean category of beliefs on $\Omega$ is defined by the following data:

(i) the objects are the subsets of $\Omega$, i.e., the elements of $\wp\Omega$,

(ii) the arrows X: A $\to$ B are the subsets X of $\Omega$ such that $X \cap A = B$,

in other words X: A $\to$ B iff $X \cap A = B$

(iii) the composite of X: A $\to$ B and Y: B $\to$ C is $X \cap Y$: A $\to$ C

(iv) the identity arrow at A is $\Omega$: A $\to$ A.

Intuitively, the Boolean category of beliefs on $\Omega$ can be explained the following way: the only admissible belief states that can be entertained are of the kind: *I believe that the answer to the question is in subset X*. The only admissible updatings are those representing the following kind of reasoning: *If I believe that the answer to the question is in subset A, and if I get an evidence which makes me believe that the answer is in subset X, then I will believe that the answer to the question is in subset $X \cap A$*.

The Bayesian category of beliefs (induced from *Bayesian machines* of [Domotor 80, p. 390] by the Meseguer-Montanari correspondence).

The Bayesian category of beliefs (on a set $\Omega$) is defined according to the following definition [Teller 73, p.218]: '*I take bayesianism to be the doctrine which maintains that (i) a set of reasonable beliefs can be represented by a probability function defined over sentences or propositions, and that (ii) reasonable changes of belief can be represented by a process called conditionalization*'.

Let $\Omega$ be a (finite or infinite) set which can be interpreted as a set of possible values for a variable, or possible answers to a question. The Bayesian category of beliefs on $\Omega$ is defined by the following data:

(i) the objects are the probability functions P: $\wp\Omega \to [0,1]$, i.e., the functions satisfying the well known Kolmogorov axioms,

(ii) the arrows X: P $\to$ Q are the subsets X of $\Omega$ such that $Q = P(. | X)$

---

[4] A preodered set is a set along with a binary relation which is reflexive and transitive.

[5] An ordered set is a set along with a binary relation which is reflexive, transitive and antisymmetric.

[6] A lattice is an ordered set in which every pair of elements has an infimum and a supremum.

(iii) the composition of X: P → Q and Y: Q → R is X∩Y: P → R
(iv) the identity arrow at P is Ω: P → P.
So, the only admissible belief states represented by this category are those represented by a probability function on Ω. The only admissible updatings are those representing the following kind of reasoning: *If my belief state (about a situation) is represented by the probability function P, and if I get an evidence which makes me believe that the answer is in subset X, then my new belief state will be represented by the conditional probability function P(. /X).*

Dempster's category of (unnormalized) beliefs
A new kind of category of beliefs was proposed by A. Dempster in the late sixties, and exposed in the seminal work of G. Shafer [76].
Let us first review the two basic notions of Dempster-Shafer theory of belief functions.
The set Ω is finite, and $\wp\Omega$ denotes its power set.
(1) A *mass distribution* m on the set Ω is any function:

$$m: \wp\Omega \to [0\ 1] \text{ such that } \sum_{X \in \wp\Omega} m(X) = 1$$

(2) The key point of the theory is provided by the so-called *Dempster's rule of combination*. It is a binary operation defined on the set of mass distributions on a set Ω: given two mass distributions $m_1$ and $m_2$, the rule provides a new mass distribution denoted by $m_1 \otimes m_2$:

$$\forall A \in \wp\Omega: m_1 \otimes m_2(A) = \sum_{X \cap Y = A} m_1(X) \cdot m_2(Y)$$

This product is in fact nothing else than the convolution product of the semi-group algebra of ($\wp\Omega, \cap$). Before I describe Dempster's category of beliefs, let us note that each subset X of Ω determines a mass distribution denoted by $1_{\{X\}}: \wp\Omega \to [0,1]$ and defined by $1_{\{X\}}(X) = 1$. I are now ready to describe what I call Dempster's category of (unnormalized) beliefs. As usual, let Ω be a finite set which can be interpreted as a set of possible values for a variable, or possible answers to a question. Dempster's category of beliefs on Ω is defined by the following data:
(i) the objects are the mass distributions m: $\wp\Omega \to [0,1]$,
(ii) the arrows e: $m_1 \to m_2$ are the mass distributions e such that: $e \otimes m_1 = m_2$,
(iii) the composite of $e_1: m_1 \to m_2$ and $e_2: m_2 \to m_3$ is $e_1 \otimes e_2: m_1 \to m_3$
(iv) the identity arrow at m is $1_{\{\Omega\}}: m \to m$.
Some comments are needed:
1. Any *mass distribution* m is bijectively represented by its *Möbius transform* also called its *belief function* $bel_m$ defined by:

$$\forall A \in \wp\Omega : bel_m(A) = \sum_{X \in \wp A - \{\emptyset\}} m(X) = \sum_{X \subseteq A, X \neq \emptyset} m(X)$$

A belief function is sometimes used instead of its mass distribution and vice versa.
2. In Dempster's unnormalized category, the belief states are represented by mathematical objects that are in fact generalized probability functions (see [Fagin, Halpern 89]).

But, what makes the situation more intricate is that the updatings (induced by evidences) are represented by the same kind of mathematical objects as belief states are. So, in this framework the phrase of Halpern and Fagin [90, p.102] receives its full meaning, namely that belief functions can be understood in *'two useful and quite different ways ... The first as a generalized probability ... The second as a way of representing evidence ...*(i.e.) *as a mapping from probability functions to probability functions'.*
3. Another point is the difference between *updating* and *combination*: An updating (transition) is an arrow from a belief state to a belief state, whereas the combination is the composition rule, operating on the arrows of the category. As stressed by Halpern and Fagin [90, p.115] : *'The key point is that updating and combining are different processes; what makes sense in one context does not necessarily make sense in the other.'* And, p.112 : *'It makes sense to think of updating a belief if we think of it as a generalized probability. On the other hand, it makes sense to combine two beliefs (using, say, Dempster' rule of combination) only if we think of the belief functions as representing evidence'.*
4. It is well known that the rule of combination of beliefs is (said to be) valid in case the *'beliefs functions to be combined are actually based on entirely distinct bodies of evidence'* [Shafer 76, p. 57].
5. It is obvious how Dempster's category of beliefs should be interpreted: the belief states are represented by mass distributions (or equivalently by belief functions) on Ω. The updatings represent the following kind of reasoning: *If my belief state is represented by the mass distribution B, and if I get an evidence – based on a body of evidence entirely distinct from the body of evidence on which is based my belief state – represented by the mass distribution E, then my new belief state will be represented by the mass distribution E⊗B.*

Dempster's category of (normalized) beliefs
The differences between this category and the unnormalized Dempster's category of beliefs are the following ones:
(i) the mass distributions are asked to satisfy m(∅)=0.
(ii) Dempster product has to be normalized, cf [Shafer 76].

Remark: there exist numerous other categories whose objects are belief functions (or mass distributions). The reader will easily define the *weak-inclusion category of beliefs* and the *strong-inclusion category* or *Yager's category of beliefs*. The main difference between the former categories and the latter ones is that the arrows of the former categories are not induced by evidence. In other words, their arrows are more *descriptive* than *operative*. The two before mentioned categories have already be somehow studied by Yager [86], Dubois and Prade [86,87b,90] and by Kruse and Schwecke [91]. According to the philosophy of category theory, and as observed by Dubois and Prade [90, p.423], these different categories *'correspond to different views of belief functions'*.

## 3 DISJUNCTIONS AND CONJUNCTIONS

A slogan for this section could be: *define the logical connectives in terms of minimum (or maximum) principles, i.e., by using universal properties.*
For example, the union and intersection (i.e., disjunction and conjunction) of two sets A and B can be defined without referring to the elements of the sets, using only the inclusion relation, in the following way:
A∪B is the set included in all sets including A and B,
A∩B is the set including all sets included in A and B.
Keeping in mind the above example here is, I believe, the essence of the conjuction and disjunction of two pieces of information:
(1) the conjunction is the piece of information *contained* in all pieces of information *containing* the two given pieces of information. More intuitively: it is the most cautious (minimal) piece of information containing the two given pieces of information.
(2) the disjunction is the piece of information *containing* all pieces of information *contained* in the two given pieces of information. More intuitively: it is the most bold (maximal) piece of information contained in the two given pieces of information.
Since Jaynes (1957), a usual approach (at least for conjunction) is the *maximum (cross-) entropy inference* or the *minimum information (gain) inference* approach, which I shall not recall here. That approach can be described as a *quantitative* approach, although only the order relation is used. Another approach, which can be described as a *qualitative* approach, is the categorial approach. Actually, the minimum information (gain) inference approach can be seen as a particular case of the categorial approach. The categorial approach – or better: the universal property approach – is the following: (i) a *piece of information* is interpreted as an object of a category (whose object can be thought of as representing belief states or information states), (ii) A is *contained* in B is interpreted as an arrow X: A → B of the category. The conjunction is then represented by a construction called the *coproduct* in the category and the disjunction is represented by another construction called the *product* in the category.

## 4 COPRODUCTS AND CONJUNCTIONS

Let us first define the simplest example of object defined by a universal property: *initial object* of a category.
**Definition**: an **initial object** of a category is an object I such that for any object X (of the category) there is a unique arrow (of the category) : I → X.
**Properties** and **examples**:
1. It can be shown very easily that all initial objects of a category are *isomorphic* (that is: if I1 and I2 are two initial objects, then there exist an arrow f:I1 → I2 and an arrow g:I2 → I1 such that f.g=$i_{I1}$ (the identity arrow at I1) and g.f=$i_{I2}$ (the identity arrow at I2).
2. The reader will verify at once that the *vacuous belief function* $1_{(\Omega)}$ is the (only) initial object in Dempster's (unnormalized or normalized) category.
3. An initial object can also be defined as a *colimit of the empty diagram* [Goldblatt 84, p. 60]. So, if one *knows nothing* (represented by the empty diagram, as suggested by Negoita [85, p. 8]), the most cautious belief state is the vacuous belief function $1_{(\Omega)}$.
4. It is easy to verify that Bayes category has no initial object.
5. The only initial object of Boole category (on Ω) is Ω.
6. The initial object of an ordered set is its minimum (if it exists).
Let us now consider two objects A and B of a category. A *coproduct* of A and B is an object A+B along with two arrows $in_A$: A → A+B and $in_B$: B → A+B, expressing how A+B is related to A and B, satisfying a particular universal property. Intuitively, A+B represents the *fusion* or *aggregation* or *integration* of A and B in the most cautious way, according to the arrows of the category being considered. Here is the definition:
**Definition**: Let us give two objects A and B (of a category)

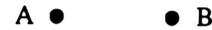

A **coproduct** of A and B is an object denoted by A+B, along with two arrows:
$in_A$: A → A+B and $in_B$: B → A+B

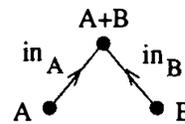

such that for every other object C along with two arrows
f: A → C and g: B → C

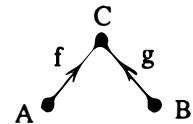

there exists a unique arrow α: A+B → C such that the following diagram *commutes*: (i.e., $in_A.\alpha=f$ and $in_B.\alpha=g$)

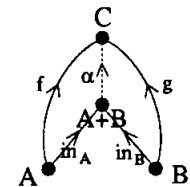

**Properties**:
1. It can be shown very easily that all coproducts of a given pair of objects are *isomorphic*.
2. [MacLane 71, p. 72-74]: up to isomorphism, the coproduct is an operation satisfying the following properties: (i) Associativity, (ii) Commutativity, (iii) Every initial object is neutral, (iv) Idempotent if the category is a preorder. To say that a category is a preorder is equivalent to saying that it has at most one arrow between an ordered pair of objects, it is also trivially equivalent to say that every diagram of the category commutes.

3. The initial object of a category can be seen to be the coproduct of ... no object!

4. A priori, there is no guarantee for a pair of objects in a category to have a coproduct. A category is said *to have binary products* if the product of any two objects exists. It is not always a trivial task to show that a category has binary products.

For classical **examples** of coproducts the reader is referred to the literature on category theory. Let us here consider the category $(\wp\Omega,\supseteq)$ which is well known to be a *Boolean lattice*. The objects are the subsets of $\Omega$, and the arrows are the ordered pairs (A,B) such that $A \supseteq B$. The reader can easily verify that the coproduct of A and B is A∩B (together with the inclusions $A \supseteq A\cap B$ and $B \supseteq A\cap B$). It is trivial to see that in any lattice the coproduct of the elements a and b is the *supremum* or *least upper bound* of a and b.

Let us now examine what happens in our Boolean category of beliefs (on the set $\Omega$). So, let us take two states A and B of the category, at first sight the coproduct of A and B should be A∩B, unfortunately the reader may verify that we get some problems in trying to get commutative diagrams which are asked by the definition of coproduct. The easiest way out of that problem is to consider that all diagrams of the category do commute! Formally, this can be done by identifying all arrows going from one state to another state – that is, by performing a *quotient* of the category. In that way we obtain a *new* category – called the *preorder* of the category – which, in the present case, is (isomorphic to) the Boolean lattice $(\wp\Omega,\supseteq)$. As the Boolean lattice $(\wp\Omega,\supseteq)$ essentially reflects the *logic* of the Boolean category, I will call the preorder of a category its **logic** (warning: this is not a standard definition).

Thus, with the preceding definition we can say that the logic of the Boolean category is a Boolean logic. We can perform the same quotient with Dempster's categories to get the *logics* of Dempster's categories.

Here are the details of the definition of the *logics* of the two Dempster's categories:

(i) the objects are the mass distributions m: $\wp\Omega \to [0,1]$,

(ii) there is only one arrow $m_1 \to m_2$ iff there exists at least one "evidence" (mass distribution) e such that: $e \otimes m_1 = m_2$, otherwise there does not exist any arrow $m_1 \to m_2$.

(iii) the composition rule and (iv) the identity arrows are then uniquely defined.

Dempster's rule of combination is normalized or not according to the category being considered.

**Definition**: Given two objects A and B of a category, the **conjunction** A∧B is the coproduct of A and B in the logic of the category.

More informally, with the vocabulary of "beliefs", the definition of conjunction is the following: the conjunction of two belief states A and B is the *least* (or more precisely: *any* least) common updated belief state of A and of B.

*'Least'* means that any other common updated belief state of A and B is an updated state of the conjunction of A and B. The conjunction is thus commutative, associative and idempotent. Moreover, in Dempster's categories the vacuous belief state is neutral for conjunction.

Because of the minimality property of the conjunction there is no need to assume *distinctness* (of the sources) of the combined beliefs.

Remark: Kruse and Schwecke [91] have succeeded in building a category, whose objects are belief functions and whose arrows are *specializations* – a generalization of Yager's inclusion [86] – in which a conjunction of A and B is Dempster's combination of A and B. At first sight this is a remarkable result! Unfortunately, because there are (too) many isomorphisms in Kruse's category, being a conjunction in that category does not characterize Dempster's rule of combination at all.

## 5 PRODUCTS AND DISJUNCTIONS

The dual notion of initial object is *terminal object*.

**Definition**: a **terminal object** of a category is an object T such that for any object X there is a unique arrow $X \to T$.

The only terminal object of the logic of the Boolean category of beliefs is the empty set $\varnothing$ which plays the role of the constant FALSE. The reader will verify at once that the belief state defined by $m(\varnothing) = 1$ is the (only) terminal object in the logic of Dempster's unnormalized category. I call that belief function the *total contradiction*, which plays, in this logic, the role of the boolean constant FALSE. Contrary to the unnormalized case, the logic of Dempster's normalized category has no terminal object (i.e., the constant FALSE is not represented in that category).

The product of two objects in a category is simply the coproduct in the *dual* category (i.e., the objects are the same but the arrows are *reversed*). Historically, *products* were first recognized which explains the word *coproduct*.

Let us now consider two objects A and B of a category. A *product* of A and B is an object A×B along with two arrows $p_A: A \times B \to A$ and $p_B: A \times B \to B$ (called the *projections*), expressing how A×B is related to A and B, and satisfying a universal property. Intuitively, an interpretation of A×B is the biggest common part of A and B according to the arrows of the category being considered. The exact definition is the following:

**Definition**: Let us give two objects A and B (of a category)

$$A \bullet \qquad \bullet B$$

A **product** of A and B is an object denoted by A×B, along with two arrows $p_A: A \times B \to A$ and $p_B: A \times B \to B$

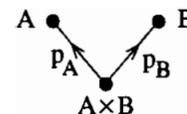

such that for every other object C (of the category) along with two arrows f: C → A and g: C → B

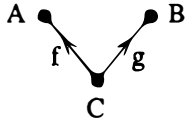

there exists a unique arrow α: C → A×B such that the following diagram *commutes*: (i.e., α.p$_A$=f and α.p$_B$=g)

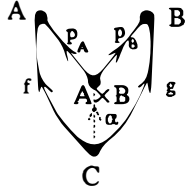

The **properties** of products are dual to those of coproducts. So, I will only remember the most important ones: up to isomorphism, the product is an operation satisfying the following properties: (i) Associativity, (ii) Commutativity, (iii) Every terminal object is neutral, (iv) Idempotent if the category is a preorder.

The reader can verify that, in the category $(\wp\Omega, \supseteq)$, the product of A and B is A∪B (together with the inclusions A∪B ⊇ A and A∪B ⊇ B). The preceding example easily extends to any lattice in which the product of the elements a and b is the *infimum* or *greatest lower bound* of a and b.

**Definition**: Given two objects A and B of a category, the **disjunction** A∨B is the product of A and B in the logic of the category.

More informally, with the vocabulary of "beliefs", the definition of disjunction is the following: the disjunction of two belief states A and B is the *most* (or more precisely: *any* most) updated belief state, such that A and B are updated states of it. '*Most*' means that the disjunction of A and B is an updated state of any state which can be updated into A and B. The disjunction is thus commutative, associative, and idempotent. Moreover, in the logic of Dempster's unnormalized category of beliefs the total contradiction is neutral for the disjunction, which fits well with the idea that the total contradiction contains all informations.

## 6   SEPARABLE BELIEF FUNCTIONS

In this last section I consider only Dempster's unnormalized category and restrict ourselves to the subcategory whose objects and arrows are separable belief functions. Let us first recall some **definitions** (see [Shafer 76]). The subset A of Ω is a *focal element* of the belief function bel$_m$: $\wp\Omega$ → [0,1] iff m(A)>0. The belief function bel$_m$: $\wp\Omega$ → [0,1] is a *simple support function* iff bel$_m$ has at most one focal element distinct from Ω. The mass distribution of any simple support function bel$_m$ *focused* on X ≠ Ω will be denoted by m$_X$. More explicitly, the mass distribution m$_X$ such that m$_X$(X) = α will be denoted by X$^\alpha$. The mass allocated to Ω by X$^\alpha$ is thus 1-α. The vacuous belief function 1$_{\{\Omega\}}$ will be represented by any expression of the from X$^0$ where X is not Ω. Note that $\varnothing^\alpha$ denotes a belief function focused on the empty set. A *separable belief function* (called *separable support function* by G. Shafer) is any belief function bel$_m$ such that:

$$m = \bigotimes_{X \in \wp\Omega\cdot\{\Omega\}} m_X$$

Using the just introduced notation we can write:

$$m = \bigotimes_{X \in \wp\Omega\cdot\{\Omega\}} X^{\alpha_X}$$

where α$_X$ = m$_X$(X)

Dempster's rule of combination gets an interesting form for separable belief functions:

**Theorem**: Given two separable belief functions,

$$\bigotimes_{X \in \wp\Omega\cdot\{\Omega\}} X^{\alpha_X} \text{ and } \bigotimes_{Y \in \wp\Omega\cdot\{\Omega\}} Y^{\beta_Y}$$

then

$$(\bigotimes_{X \in \wp\Omega\cdot\{\Omega\}} X^{\alpha_X}) \otimes (\bigotimes_{Y \in \wp\Omega\cdot\{\Omega\}} Y^{\beta_Y}) = \bigotimes_{X \in \wp\Omega\cdot\{\Omega\}} X^{\alpha_X + \beta_X - \alpha_X \beta_X}$$

The next theorem states that the conjunction (∧) and disjunction (∨) of two beliefs in the category of separable belief functions always exist and are given by the following formulas:

**Theorem**[7]: Given two separable belief functions,

$$\bigotimes_{X \in \wp\Omega\cdot\{\Omega\}} X^{\alpha_X} \text{ and } \bigotimes_{Y \in \wp\Omega\cdot\{\Omega\}} Y^{\beta_Y}$$

then

$$(\bigotimes_{X \in \wp\Omega\cdot\{\Omega\}} X^{\alpha_X}) \wedge (\bigotimes_{Y \in \wp\Omega\cdot\{\Omega\}} Y^{\beta_Y}) = \bigotimes_{X \in \wp\Omega\cdot\{\Omega\}} X^{max(\alpha_X, \beta_X)}$$

$$(\bigotimes_{X \in \wp\Omega\cdot\{\Omega\}} X^{\alpha_X}) \vee (\bigotimes_{Y \in \wp\Omega\cdot\{\Omega\}} Y^{\beta_Y}) = \bigotimes_{X \in \wp\Omega\cdot\{\Omega\}} X^{min(\alpha_X, \beta_X)}$$

It is interesting to note that the operations induced on the "exponents" by Dempster's rule of combination and the conjunction are T-conorms[8], and the operation induced by the disjunction is a T-norm[9]. The link between T-(co)norms and fuzzy logical connectives has been recognized since a long time ago. This was first recognized by U. Höhle in the late seventies. It can be shown very easily that T-(co)norms is a weakened form of the notion of categorial (co)products in the category [0,1],≤.

## 7   CONCLUSIONS

The key point expressed in the present paper is that Dempster's rule of combination is less a conjunctive rule

---

[7] These two formulas - without categorial content - have been brought to my attention by Philippe Smets.

[8] A T-conorm is an operation *: [0,1]×[0,1] → [0,1] which is commutative, associative, for which 0 is neutral, monotonic increasing (for the usual order relation ≤ on [0,1] and the product order on [0,1]×[0,1]), and 1*1=1.

[9] A T-norm is an operation *: [0,1]×[0,1] → [0,1] which is commutative, associative, for which 1 is neutral, monotonic increasing, and 0*0=0.

(of belief states) than the composition rule of a category (composition of evidence). Such a point of view has lead to conjunction and a disjunction for belief functions. Explicit formulas have been given for separable belief functions.

The former key point may also explain a difference between fuzzy logic and Dempster-Shafer theory as follows. In fuzzy logic there exist numerous "combination rules" modelling the logical connectives. People are trying to unify them all by finding a category out of which all these connectives can emerge naturally. Whereas in Dempster-Shafer theory there exists only one combination rule, and people are trying to discover a logic (cf. e.g. [Dubois, Prade 86]) compatible with the combination rule.

## Acknowledgments

Philippe Smets once suggested to me to find out an 'idempotent Dempster's rule of combination' to be used in non-independent situations. The ideas contained in the present paper have grown out of that suggestion for which I am very grateful. I thank Bruno Marchal for his constant encouragement to use a categorial viewpoint and for uncountable many stimulating discussions. I gratefully acknowledge Yen-Teh Hsia and Alessandro Saffiotti for countable many and very fruitful discussions. I am deeply greatful to Yen-Teh for a very careful reading of the paper and for his many comments about it.